# Application of Deep Interpolation Network for Clustering of Physiologic Time Series

Yanjun Li [4], Yuanfang Ren, PhD[1, 5], Tyler J. Loftus, MD[2,5], Shounak Datta, PhD[1, 5], ,M. Ruppert, BS[1, 5], Ziyuan Guan, MS[1,5], Dapeng Wu, PhD[4], Parisa Rashidi, PhD[3,5], Tezcan Ozrazgat-Baslanti, PhD[1,5,6*], Azra Bihorac, MD[3,5,,6*]

[1]Department of Medicine, Division of Nephrology, Hypertension, and Renal Transplantation, University of Florida, Gainesville, FL.

[2]Department of Surgery, University of Florida, Gainesville, FL.

[3]J. Crayton Pruitt Family Department of Biomedical Engineering, University of Florida, Gainesville, FL.

[4] NSF Center for Big Learning, University of Florida, Gainesville, FL.

[5]Precision and Intelligent Systems in Medicine (PrismaP), University of Florida, Gainesville, FL

[6]Sepsis and Critical Illness Research Center, University of Florida, Gainesville, FL.

*These authors contributed equally.

**Correspondence to:**

Azra Bihorac MD MS, Department of Medicine, Division of Nephrology, Hypertension, and Renal Transplantation, PO Box 100224, Gainesville, FL 32610-0254. Phone: (352) 273-9009; Fax: (352) 392-5465; Email: abihorac@ufl.edu

**Funding:** A.B., T.O.B., and P.R. were supported by R01 GM110240 from the National Institute of General Medical Sciences. A.B. and T.O.B. were supported by Sepsis and Critical Illness Research Center Award P50 GM-111152 from the National Institute of General Medical Sciences. T.O.B. has received grant that was supported by the National Center For Advancing Translational Sciences of the National Institutes of Health under Award Number UL1TR001427 and received grant from Gatorade Trust (127900), University of Florida. This work was supported in part by the NIH/NCATS Clinical and Translational Sciences Award to the University of Florida UL1 TR000064. The content is solely the responsibility of the authors and does not necessarily

**Abstract**

**Background:** During the early stages of hospital admission, clinicians must use limited information to make diagnostic and treatment decisions as patient acuity evolves. However, it is common that the time series vital sign information from patients to be both sparse and irregularly collected, which poses a significant challenge for machine / deep learning techniques to analyze and facilitate the clinicians to improve the human health outcome. To deal with this problem, We propose a novel deep interpolation network to extract latent representations from sparse and irregularly sampled time-series vital signs measured within six hours of hospital admission

.

**Methods:** We created a single-center longitudinal dataset of electronic health record data for all (n=75,762) adult patient admissions to a tertiary care center lasting six hours or longer, using 55% of the dataset for training, 23% for validation, and 22% for testing. All raw time series within six hours of hospital admission were extracted for six vital signs (systolic blood pressure, diastolic blood pressure, heart rate, temperature, blood oxygen saturation, and respiratory rate). A deep interpolation network is proposed to learn from such irregular and sparse multivariate time series data to extract the fixed low-dimensional latent patterns. Based on the extracted pattern, we use k-means clustering algorithm to clusters the patient admissions resulting into 7 clusters.

**Findings:** Training, validation, and testing cohorts had similar age (55-57 years), sex (55% female), and admission vital signs. Seven distinct clusters were identified. Moreover, clustering the patients based on the features generated by our deep interpolation network achieves much better results than working on the hand-crafted features, in terms of the standard clustering internal measurements, such as Silhouette score and Davie-Brown Index.

**Interpretation:** In a heterogeneous cohort of hospitalized patients, a deep interpolation network extracted representations from vital sign data measured within six hours of hospital admission. This approach may have important implications for clinical decision-support under time constraints and uncertainty.

**Introduction**

In the United States, there are more than 36 million hospital admissions and seven thousand in-hospital deaths each year, nearly one quarter of which may be preventable.[1-4] In the early stages of hospital admission, misdiagnosis and under-triage of high-risk patients to general hospital wards appear to be major sources of preventable harm.[5, 6] Physicians are often blind to their own errors in judgement unless feedback is provided by post-mortem examinations, of which 10-15% reveal major diagnostic errors.[7-9]

During the early stages of hospital admission, clinicians must use limited information to try to understand underlying disease pathology and recommend and order diagnostic tests, treatments, and triage destinations for patients.[7-9] To perform this task, clinicians analyze vital signs, which represent essential physiological processes. Values and trends in vital signs can indicate the appropriateness of frequent monitoring in an intensive care unit (ICU) versus low-intensity care on a general hospital ward. [10-14] Early values and trends in vital signs may also make it possible to identify unique physiological signatures associated with distinct patient phenotypes and clinical outcomes. Clustering analyses using vital signs and other clinical variables have identified clinically meaningful sepsis and heart failure phenotypes, but this approach has not been reported among broad, heterogeneous cohorts incorporating all hospitalized patients.[15, 16]

One of the major challenges in clustering on time-series vital signs is that they are irregularly sampled. In this paper, we propose a novel deep interpolation network to extract latent representations from sparse and irregularly sampled time-series vital signs measured within six hours of hospital admission. Based on the extracted features, clustering analyses identified seven distinct patient phenotypes.

**Methods**

**Data Source and Participants**

This project was approved by the University of Florida (UF) institutional review board under a waiver of informed consent and with authorization under the Health Insurance Portability and Accountability Act. Using the UF Health (UFH) Integrated Data Repository as Honest Broker, we created a longitudinal dataset from electronic health records of all adults (age ≥18 years) admitted to the 800-bed academic hospital at UFH between June 1, 2014 and April 1, 2016. The dataset includes structured and unstructured clinical data, demographic information, vital signs, laboratory values, medications, diagnoses, and procedures. Patients with less than six hours of admission and those completely missing at least two of the six vital sign measurements (systolic and diastolic blood pressure, heart rate, respiratory rate, temperature, and peripheral capillary oxygen saturation) in the first six hours of admission were excluded from the analysis (Figure 1). The final cohort consisted of 75,762 hospital admissions for 43,598 patients.

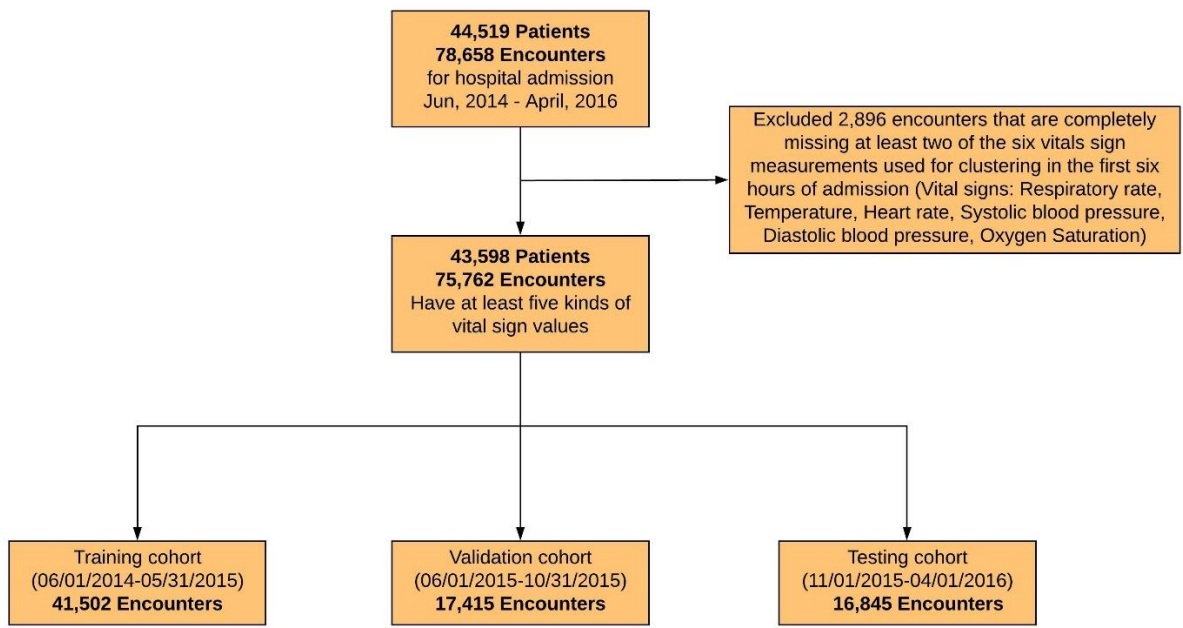

**Figure 1.** Cohort selection and exclusion criteria

**Study Design**

We non-randomly split the dataset by admission dates into three cohorts: training (admissions between June 1, 2014 and May 31, 2015, n = 41,502, 55% of all admissions), validation (admissions between June 1, 2015 and October 31, 2015, n = 17,415, 22% of all admissions), and testing (admissions between November 1, 2015 and April 1, 2016, n = 16,845, 23% of all admissions). To determine acute illness phenotypes using early physiologic signatures, we applied unsupervised clustering methods to the repeated measurements of six vital signs available within the first six hours of hospital admission in the training cohort.

**Development of Physiologic Signatures Using Vital Signs Time Series**

We selected six vital signs, ubiquitously and repeatedly measured during hospitalization, representing unique physiologic responses - systolic and diastolic blood pressure, heart rate, respiratory rate, temperature, and peripheral capillary oxygen saturation. For each vital sign, raw time series (all available measurements with time stamp in EHR) within the first six hours of hospital admission were processed to remove outliers and assess distributions, missingness, and correlation.

**Demographic, Diagnostic and Biological Correlates and Clinical Outcomes**

For each admission we extracted demographic variables (e.g., age, sex, Elixhauser comorbidities), all diagnostic and procedural codes, and several serum biomarkers routinely measured at baseline hospital admission, broadly categorized under the domains of inflammatory, endothelial, coagulation, and vital organ function.

The primary outcomes were thirty-day and three-year mortality. Other outcomes included for exploratory analyses included hospital complications (acute kidney injury (AKI), sepsis, cardiovascular complications), intensive care unit admission and duration, mechanical ventilation and duration, and discharge disposition.

**Analytic methods**

**Deep Interpolation Network**

In this section, we describe our proposed Deep Interpolation Network (DIN) for clustering the patients based on their vital sign data during the early stages of hospital admission. Using the raw sparse and irregularly sampled time series vital sign as the input, DIN can automatically extract a unified and abstract representation of the entire time-series data of an encounter via an end-to-end unsupervised manner. The overall network architecture consists of four main compounds: Interpolation model, Seq2Seq model, Re-interpolation model and Clustering model.

Figure 1 provides the schematic representation of the DIN architecture and feature learning process. For the interpolation model, we adopt the recent work [28] to first interpolate the raw time-series vital sign data to a regularly sampled meta-representation with pre-defined reference time points. Then we feed the interpolated time-series data into a Seq2Seq model with GRU [30] layers for feature embedding and extracting a unified context vector lying in the low-dimensional feature space by the encoder. The context vector contains the global time-series information and is further used by any standard clustering methods for the patient clustering. The decoder in the Seq2Seq model learns from the context vector and outputs the time-series data with the same length of the Seq2Seq model's input. Then, we deploy a radial basis function network-based model to re-interpolate the fixed-length output to the raw irregular time points for reconstructing the raw vital signs data at corresponding time points. The full feature extraction model is end-to-end trained by minimizing the reconstruction loss measured with mean square error. We describe the components of the DIN in detail in the following subsections.

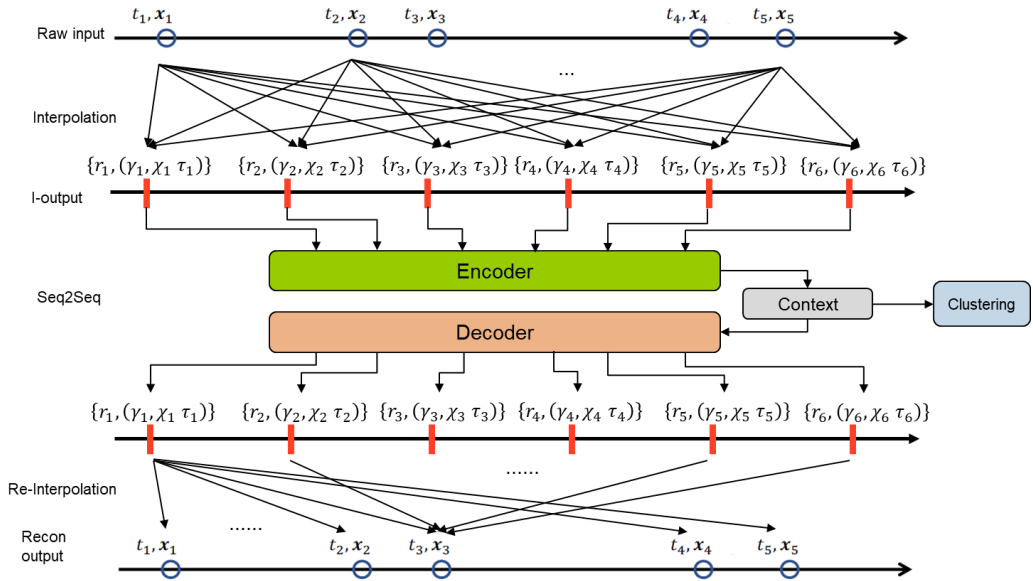

**Figure 2.** Schematic representation of Deep Interpolation Network for patient clustering. For brevity, we drop some connections in the interpolation and re-interpolation process. Indeed, values at each raw time point contribute to all the referenced time points in the interpolation phase and vice versa in the re-interpolation phase.

**Interpolation Model**

It is common that the time series vital sign data in electronic health records to be both sparse and irregularly sampled, which means large and irregular intervals widely exist between the data observation time points. Such sparsity and irregularity pose a significant challenge for machine / deep learning techniques to analyze the crucial vital sign data for improving the human health outcome. To deal with this problem, we adopt the network proposed by Shukla and Marlin [28] first to interpolate the raw time-series data to a regularly sampled meta-representation with pre-defined reference time points.

In our study, we utilize six vital signs multivariate time series data, e.g., two kinds of blood pressure (systolic and diastolic), heart rate, temperature, Spo2, and respiratory rate. Take one variable out of six as an example. For one patient, the raw time-series data is denoted as $e = \{(t_i, x_i) | i = 1, \ldots, I\}$, where $I$ represents the total number of observations, $t_i$ is the time point, and $x_i$ is the corresponding observed value. The time intervals between adjacent observation time points vary a lot. The interpolation model can map irregular $e$ value to the regular time series data which is defined at the $T$ reference time points $r = [r_1; \ldots; r_T]$ with evenly spaced interval.

The interpolation model consists of two layers, where the first layer separately performs the interpolation for each variable, and the second layer aggregates the information across all the studied variables. The model generates three different channel groups at each reference time point, which respectively represents smooth trends $\chi$, short time-scale transients $\tau$, and local observation frequencies $\lambda$. The interpolation model enables the single observation data point to be considered by all the reference time points and allows for the information to be shared across multiple variables. For more detailed interpolation mathematic denotation, the reader is referred to [28].

**Seq2Seq Model**

With the interpolated time-series data as the input, we develop a Seq2Seq model to learn its low-dimensional representation, which can embed the contextual information over the full timeline. Seq2Seq model is a method of the encoder-decoder framework that maps an input of sequence to an output of sequence, and it is broadly used in machine translation, text summarization, conversational modeling, and some other tasks. With a single layer GRU network [30] as the encoder, the input sequence is encoded to a fixed-length contextual vector $h_T$, which is the hidden state of the last time step. The hidden state of GRU updating mechanism, illustrated in the following equation, ensures that every internal hidden node state will be calculated by the previous state $h_{i-1}$ and current time step input $(\chi_i, \tau_i, \lambda_i)$.

$$h_i = GRU_{Enc}((\chi_i\ \tau_i,\ \lambda_i),\ h_{i-1})$$

A single-layer GRU network is also used for a decoder. At each time step, the decoder updates its current hidden state $s_t$ with the concatenated features incorporating the previous decoded output $o_{t-1}$ and global context vector $h_T$ as the input:

$$s_t = GRU_{Dec}([o_{t-1};\ h_T], s_{t-1})$$

**Re-Interpolation Model**

To unsupervised learn the useful representation, a common strategy is to build an autoencoder learning framework by reconstructing the input itself from the extracted bottleneck representation. Therefore, on top of the Seq2Seq model, we develop a re-interpolation network to map the output with the evenly spaced intervals to the raw irregular time points. Similar to the interpolation model, the transformation is also based on a radial basis function network. Our re-interpolation model allows the embedded values at every reference time point to make a continuous contribution to reconstructed values at all the raw time points, but the contribution weight is exponentially decayed in terms of the distance between the referenced time point $r_i$ and target time point $t_j$:

$$w(r_i, t_j, \theta) = exp(-\theta(r_i - t_j)^2)$$

where $\theta$ is learnable network parameters.

After the re-interpolation, we can easily calculate the mean square error at every input time point and minimizing this reconstruction loss is served as the learning objective of the full DIN model. It is worth noting that the interpolation, Seq2Seq, and re-interpolation models in the DIN are jointly optimized. Compared with the work [28], it effectively improves the model learning capacity and allows the clustering representation to contain more global information across the full timeline. After the model training, we also visualize the reconstruction performance of the test cohort to verify our model learning capacity.

**Clustering Model**

Taking the low-dimensional feature generated by our DIN, any standard clustering algorithm can be used to derive phenotypes in the cohort. In our case, we apply the centroid-based classical k-means clustering to the features derived from time series of six vital signs measured within six hours of hospital admission for each encounter. We determine the optimal number of phenotypes using the broadly adopted "Gap-statistic" algorithm.

Patterns of vital signs were visualized by two different methods: (1) line plots with a 95% confidence interval band, which illustrate the average value of variables used for clustering across phenotypes over time, (2) t-distribution stochastic neighbor embedding plots, which illustrate multidimensional data in two dimensions.

**Method Comparison**

To demonstrate the effectiveness of learnt representation by our deep interpolation network, we also develop a baseline approach to generate the representation with handcrafted feature engineering. For this baseline approach, raw time series were resampled to an hourly frequency, taking the mean value when multiple measurements existed during the same one-hour window. Following resampling, the missing values were replaced using first forward-propagating previous values and then back-propagating posterior values. For all remaining missing values, which is due to having no measurements during the hospitalization in the plausible range for a variable, median values of corresponding variables in the training cohort were imputed. Thus, for each admission we had six values for each of the six vital signs, producing 36 clustering input features. Based on the two sets of features generated by different interpolation methods, we respectively run the k-means algorithms with the same hyper-parameter setting to cluster the patients. To compare the clustering results, we use two different internal clustering evaluation metrics: Davies-Bouldin Index (DBI) [31] and Silhouette score [32].

For the cohort set $E = \{e_1, e_2, ..., e_m\}$, where $e_i$ represents the time series data of $i$th encounter and $m$ is the number of the encounters. Suppose the clustering results are denoted as $C = \{C_1, C_2, ..., C_k\}$, we can respectively obtain the average distance $avg(C)$ within the cluster $C$, the farthest distance $diam(C)$ within the cluster $C$, the minimum distance $d_{min}(C_i, C_j)$ between cluster $C_i$ and $C_j$, and cluster center distance $d_{cen}(C_i, C_j)$ between cluster $C_i$ and $C_j$ using the following equations:

$$avg(C) = \frac{2}{|C|(|C|-1)} \sum_{1<i<j\leq|C|} dist(x_i, x_j)$$

$$diam(C) = \max_{1<i<j\leq|C|} dist(x_i, x_j)$$

$$d_{min}(C_i, C_j) = \min_{x_i \in C_i, x_j \in C_j} dist(x_i, x_j)$$

$$d_{cen}(C_i, C_j) = dist(\mu_i, \mu_j)$$

where $dist(x_i, x_j)$ represents the distance (e.g. Euclidean distance) between two encounters $x_i$ and $x_j$ in the clustering space; $\mu$ denotes the center of the cluster $\mu = \frac{1}{|C|}\sum_{1\leq i\leq|C|} x_i$.

Based on these distances, DBI can be calculated as follows:

$$\text{DBI} = \frac{1}{k}\sum_{i=1}^{k} \max_{j\neq i}\left(\frac{avg(C_i) + avg(C_j)}{d_{cen}(\mu_i, \mu_j)}\right)$$

If the clustering method can provide a lower DBI value, we can claim that this method achieves better clustering results than the others.

For the Silhouettes score, it measures how similar an object is to its own cluster (cohesion) compared to other clusters (separation). For each patient/encounter $e_i$, the silhouette score is given by

$$s(i) = \frac{b(i) - a(i)}{max\{a(i), b(i)\}}, if\ |C_i| > 1$$

where $a(i) = \frac{1}{|C_i|-1}\sum_{j\in C_i, i\neq j} d(i,j)$ denotes the average distance to all other patients within the same cluster and $b(i) = \min_{k\neq i} \frac{1}{|C_k|}\sum_{j\in C_k} d(i,j)$ denotes the smallest mean distance of i to all points in any other cluster.

Silhouette analysis is broadly used to measure how close each point in one cluster is to points in the neighboring clusters, that is, the separation distance between the resulting clusters. This measure has a range of [-1, 1]. Silhouette coefficients near +1 indicate that the sample is far away from the neighboring clusters. A value of 0 indicates that the sample is on or very close to the decision boundary between two neighboring clusters. Negative values indicate that those samples might have been assigned to the wrong cluster.

**Reproducibility**

To determine reproducibility in external data, we used the validation cohort and rederived groups using *k*-means clustering. Patterns of clinical variables were again assessed by line plots and t-distribution stochastic neighbor embedding plots. Patterns of clinical variables across clusters were compared between training and validation cohorts.

To determine the reproducibility of the phenotypes in the testing cohort, we predicted phenotypes using the clinical characteristics of typical cluster members in the derivation cohort. Predictions arose from the Euclidean distance from each patient to the centroid of each phenotype.

Consider the *i*th subject with *p* features. We represent it as $X_i = [x_1, x_2, \cdots, x_p]$. We denote the mean of the *k*th phenotype with $\mu_k = [\mu_{k1}, \mu_{k2}, \cdots, \mu_{kp}]$ and represent it as the center of the phenotype. Thus, we calculate the Euclidean distance of the *i*th admission to the center of the *k*th phenotype, $d_{i,k}$ as:

$$d_{i,k} = \sqrt{\sum_{j=1}^{p}(x_{ij} - \mu_{kj})^2}$$

We calculated distances of all admissions to all phenotype centroids and assigned each admission to its nearest phenotype.

Following the phenotype prediction, we performed qualitative analyses of clusters by examining the clinical characteristics of the predicted phenotype groups. We determined the association of the phenotypes with demographics, comorbidities, primary diagnosis groups, procedures, and several biomarkers. We presented continuous variables as mean (SD) and median values with interquartile ranges; we presented categorical variables as frequencies and percentages. We compared clinical characteristics and outcomes between the clusters using χ2 test for categorical variables and using analysis of variance and the Kruskal-Wallis test for continuous variables as appropriate. Overall survival of each cluster was illustrated using Kaplan–Meier curves. Differences in survival among clusters were tested using the log-rank test. Unadjusted and adjusted hazard ratios (HR) for each cluster compared with Cluster 1 was obtained by using Cox proportional-hazards regression, adjusting for age group ($\geq$ 65 vs. <65), race, comorbidity index ($\geq$ 3 vs <3), and SOFA score (0-1, 2-4, $\geq$ 5). We adjusted for the family-wise error rate due to multiple comparisons by adjusting $p$ values with the Bonferroni correction. To ensure that phenotypes did not overlap with traditional clinical grouping we compared them with the worst Sequential Organ Failure Assessment (SOFA) score within 24 hours of admission using alluvial plots. To further explore organ failure across six organ systems, chord diagrams were created. Analyses were performed with Python version 3.6.

**Results**

**Patients**

Three cohorts (i.e., training, validation, and testing) were used to build and validate the model. All cohorts had similar clinical characteristics. The average age of patients was 54 and sex were equally distributed.

**Reconstruction Performance of DIN**

To unsupervised learn the useful representation, a common strategy is to build an autoencoder learning framework by reconstructing the input itself from the extracted bottleneck representation. Therefore, on top of the Seq2Seq model, we develop a re-interpolation network to map the output with the evenly spaced intervals to the raw irregular time points.. Many past researches have used similar strategies to avoid the problem of trivially memorizing the input data without learning useful structure. The learning target is to reconstruct irregular vital sign data and we set the mean square error loss as our training loss and visualize the reconstruction results of the test cohort in Figure 3.

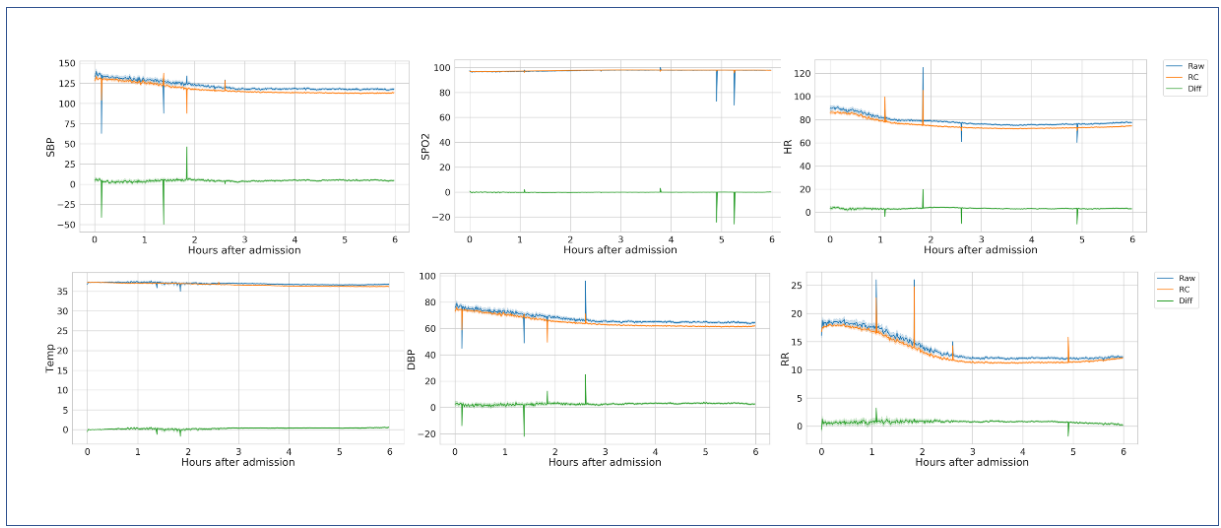

**Figure 3** Reconstruction performance of different vital signs on the testing cohort. Raw and RC represent the raw and reconstructed vital signs, respectively. Diff represents differences between the counterpart raw and reconstructed values.

**Derivation of Clinical Phenotypes**

In the training cohort, the k means clustering models found that the optimal number of phenotype clusters was seven. We use the "Gap-Statistic" method to select the optimal number of clusters by fitting the model with a wide range of values for K. According to the "Gap-Statistic" results of the validation cohort shown in the Figure 4**Error! Reference source not found.**, we finalize the optimal cluster number as seven.

With the optimal cluster number, we run the k-means algorithm on the training cohort and generate the cluster labels for the patients in the testing cohort.

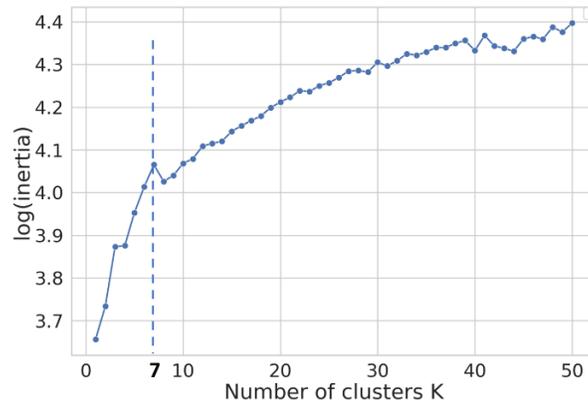

**Figure 4** The gap-statistic value with the different numbers of clusters measured on the validation cohort. The optimal k value is 7.

**Comparison between Our Interpolation Method and the Baseline Approach**

Developed on the same training cohort, our method uses the novel deep learning model to interpolate the hidden representations, whereas the baseline method uses the hand-crafted feature engineering. Then same k-means algorithm is applied on the two sets of generated features and is validated on the same validation cohort. The evaluation metrics are above described clustering internal indexes, i.e., Sihouette and DBI scores. Based on the results shown in the Table 1, we find that our DIN method achieves much better results than the forward feeding method.

**Table 1** Comparison of baseline and proposed methods (with the same cluster number) on Silhouette and DBI scores

| Metrics | Forward Feeding Method (clusters=7) | | | DIN Method (clusters=7) | | |
|---|---|---|---|---|---|---|
| Subsets | Training | Validation | Testing | Training | Validation | Testing |
| Silhouette | 0.094 | 0.095 | 0.094 | 0.208 | 0.199 | 0.211 |
| DBI | 2.097 | 2.085 | 2.082 | 1.228 | 1.127 | 1.222 |

Abbreviation. DBI, Davies-Bouldin Index.

**Analysis on Clustering Results Generated by DIN**

First, phenotypes were visualized using t-distributed stochastic embedding (t-SNE) to group admissions with the same phenotype. As shown in Figure 5, seven distinct clusters are partitioned.

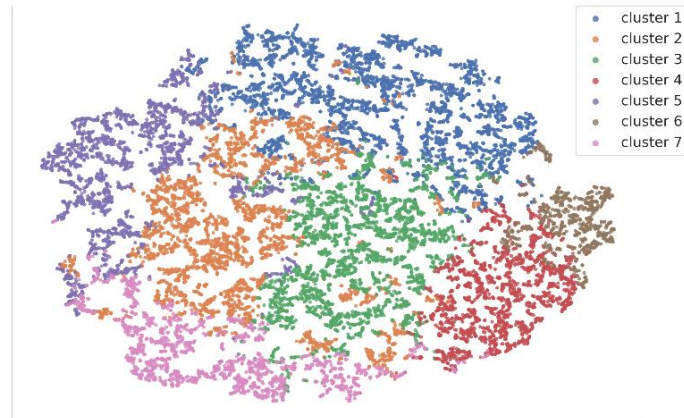

**Figure 5** t-SNE plot for the reduced 2-D feature space of the patients in the testing cohort.

Figure 6 demonstrates the time-series distribution of the six vital signs across different phenotypes for the testing cohort in the line-plot.

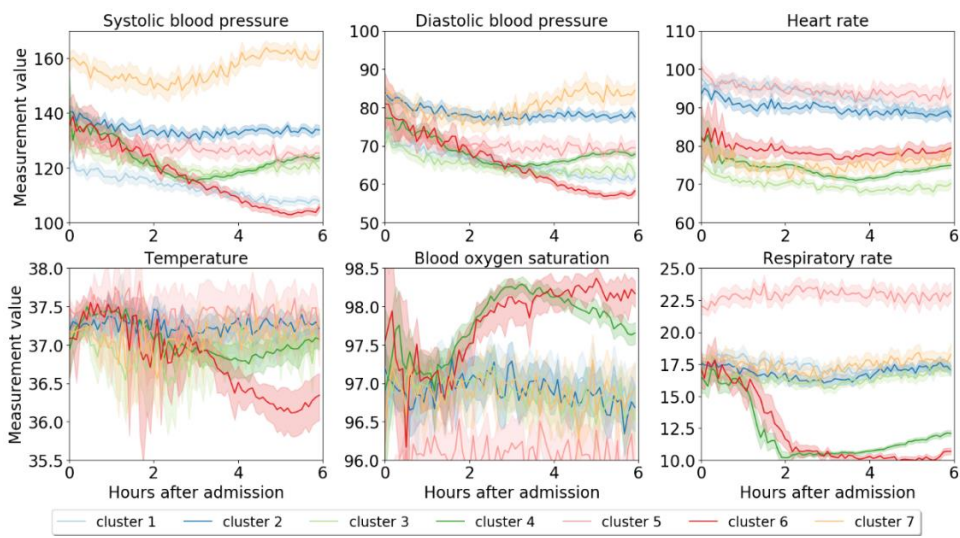

**Figure 6** Test cohort clusters had unique distributions of vital signs during the first six hours of admission

**Characteristics of Phenotypes with Biomarker Profile**

In a heterogenous cohort of all adult patients admitted to hospital for six hours or more, we identified seven clusters of patients that had unique demographic factors, disease processes, and short- and long-term clinical outcomes. Notably, these clusters were identified using raw time series vital sign data within six hours of admission, which is available for virtually any patient in any health care setting. A deep learning model was used to interpolate hidden representations, offering better clustering than a baseline forward-feeding method using hand-crafted feature engineering. This performance advantage was manifest as lower DBI and higher Silhouette scores. The clusters identified by DIN methods could be clinically relevant for prognostic and clinical decision-making tasks by identifying unique physiological signatures associated with distinct patient phenotypes and clinical outcomes.

**Table 2** Clinical and Outcome Characteristics by Phenotypes

| Variables | Total | Cluster 1 | Cluster 2 | Cluster 3 | Cluster 4 | Cluster 5 | Cluster 6 | Cluster 7 |
|---|---|---|---|---|---|---|---|---|
| Number of encounters, n (%) | 16845 | 3825 (23) | 3574 (21) | 2887 (17) | 1833 (11) | 2206 (13) | 783 (5) | 1737 (10) |
| Number of patients, n (%) | 12559 | 3262 (26) | 3114 (25) | 2571 (20) | 1755 (14) | 1947 (16) | 779 (6) | 1558 (12) |
| **Demographics** | | | | | | | | |
| Age, years, median (25th, 75th) | 57 (40, 69) | 50 (31, 64) | 52 (36, 64) | 60 (43, 72) | 61 (49, 70) | 60 (45, 72) | 59 (46, 68) | 64 (52, 76) |
| Age, mean (SD) | 54.7 (18.8) | 48.4 (18.9) | 51.1 (17.9) | 57.0 (19.6) | 58.4 (15.8) | 57.9 (18.9) | 56.1 (17.0) | 63.2 (17.0) |
| Female sex, n (%) | 9205 (55) | 2366 (62) | 1866 (52) | 1567 (54) | 932 (51) | 1165 (53) | 418 (53) | 891 (51) |
| Race, n (%) | | | | | | | | |
| White | 11854 (70) | 2697 (71) | 2361 (66) | 2160 (75) | 1431 (78) | 1506 (68) | 614 (78) | 1085 (62) |
| African American | 3845 (23) | 813 (21) | 995 (28) | 550 (19) | 241 (13) | 579 (26) | 100 (13) | 567 (33) |
| Others | 1146 (7) | 315 (8) | 218 (6) | 177 (6) | 161 (9) | 121 (5) | 69 (9) | 85 (5) |
| **Comorbidities** | | | | | | | | |

| Variables | Total | Cluster 1 | Cluster 2 | Cluster 3 | Cluster 4 | Cluster 5 | Cluster 6 | Cluster 7 |
|---|---|---|---|---|---|---|---|---|
| Charlson Comorbidity Index, median (25th, 75th) | 3 (2, 6) | 3 (2, 6) | 3 (2, 6) | 3 (2, 6) | 3 (2, 6) | 3 (2, 6) | 3 (2, 6) | 3 (2, 6) |
| Hypertension, n (%) | 8468 (50) | 1889 (49) | 1815 (51) | 1489 (52) | 917 (50) | 1096 (50) | 403 (51) | 859 (49) |
| Cardiovascular disease, n (%) | 4702 (28) | 1083 (28) | 995 (28) | 802 (28) | 492 (27) | 651 (30) | 216 (28) | 463 (27) |
| Diabetes mellitus, n (%) | 3945 (23) | 853 (22) | 833 (23) | 666 (23) | 445 (24) | 535 (24) | 197 (25) | 416 (24) |
| Chronic kidney disease, n (%) | 2892 (17) | 573 (15) | 610 (17) | 524 (18) | 193 (11) | 486 (22) | 56 (7) | 450 (26) |
| Moderate/Severe (>= Stage 3), n (%) | 1277 (8) | 217 (6) | 225 (6) | 268 (9) | 73 (4) | 225 (10) | 24 (3) | 245 (14) |
| Preadmission estimated glomerular filtration rate (mL/min per 1.73 m2), median (25th, 75th) | 96 (78, 114) | 103 (85, 122) | 101 (83, 118) | 93 (75, 109) | 94 (80, 107) | 94 (73, 111) | 94 (81, 109) | 83 (48, 101) |
| End stage kidney disease, n (%) | 914 (5) | 196 (5) | 181 (5) | 123 (4) | 62 (3) | 137 (6) | 32 (4) | 183 (11) |
| Cerebrovascular disease, n (%) | 2995 (18) | 665 (17) | 682 (19) | 508 (18) | 275 (15) | 411 (19) | 130 (17) | 324 (19) |

| Variables | Total | Cluster 1 | Cluster 2 | Cluster 3 | Cluster 4 | Cluster 5 | Cluster 6 | Cluster 7 |
|---|---|---|---|---|---|---|---|---|
| History of cancer, n (%) | 3036 (18) | 694 (18) | 663 (19) | 501 (17) | 355 (19) | 367 (17) | 136 (17) | 320 (18) |
| **Admission characteristics of patients** | | | | | | | | |
| **Admitting type, n (%)** | | | | | | | | |
| Emergent | 11920 (71) | 2852 (75) | 2857 (80) | 2290 (79) | 263 (14) | 1993 (90) | 119 (15) | 1546 (89) |
| Routine elective | 4556 (27) | 871 (23) | 630 (18) | 539 (19) | 1548 (84) | 152 (7) | **648 (83)** | 168 (10) |
| Trauma center | 369 (2) | 102 (3) | 87 (2) | 58 (2) | 22 (1) | 61 (3) | 16 (2) | 23 (1) |
| Transfer from another hospital, n (%) | 2859 (17) | 716 (19) | 648 (18) | 561 (19) | 74 (4) | 483 (22) | 53 (7) | 324 (19) |
| Admission to surgical services, n (%) | 4780 (28) | 681 (18) | 668 (19) | 661 (23) | 1529 (83) | 290 (13) | 655 (84) | 296 (17) |
| Surgery on admission day, n (%) | 3551 (21) | 412 (11) | 335 (9) | 312 (11) | 1558 (85) | 93 (4) | 693 (89) | 148 (9) |
| Surgery at any time, n (%) | 4718 (28) | 716 (19) | 626 (18) | 560 (19) | 1588 (87) | 246 (11) | 705 (90) | 277 (16) |
| Type of surgery, n (%) | | | | | | | | |
| Cardiothoracic surgery | 351 (7) | 50 (7) | 26 (4) | 40 (7) | 76 (5) | **40 (16)** | **104 (15)** | 15 (5) |

| Variables | Total | Cluster 1 | Cluster 2 | Cluster 3 | Cluster 4 | Cluster 5 | Cluster 6 | Cluster 7 |
|---|---|---|---|---|---|---|---|---|
| Non-cardiac general surgery | 1344 (28) | 206 (29) | 220 (35) | 170 (30) | 419 (26) | 76 (31) | 142 (20) | 111 (40) |
| Transplant surgery | 111 (2) | 10 (1) | 14 (2) | 11 (2) | 45 (3) | 5 (2) | 17 (2) | 9 (3) |
| Acute care and burn surgery | 373 (8) | 87 (12) | 88 (14) | 61 (11) | 48 (3) | 31 (13) | 24 (3) | 34 (12) |
| Vascular surgery | 471 (10) | 76 (11) | 68 (11) | 66 (12) | 136 (9) | 29 (12) | 47 (7) | 49 (18) |
| General gastrointestinal surgery | 227 (5) | 17 (2) | 27 (4) | 20 (4) | 119 (7) | 2 (1) | 28 (4) | 14 (5) |
| General oncology surgery | 162 (3) | 16 (2) | 23 (4) | 12 (2) | 71 (4) | 9 (4) | 26 (4) | 5 (2) |
| Neurologic surgery | 642 (14) | 47 (7) | 88 (14) | 90 (16) | 254 (16) | 37 (15) | 78 (11) | 48 (17) |
| Specialty surgery | 2202 (47) | 376 (53) | 279 (45) | 244 (44) | 779 (49) | 90 (37) | 341 (48) | 93 (34) |
| Ear nose throat | 276 (6) | 27 (4) | 24 (4) | 19 (3) | 125 (8) | 13 (5) | 52 (7) | 16 (6) |
| Orthopedics surgery | 1099 (23) | 162 (23) | 123 (20) | 130 (23) | 370 (23) | 49 (20) | 212 (30) | 53 (19) |

| Variables | Total | Cluster 1 | Cluster 2 | Cluster 3 | Cluster 4 | Cluster 5 | Cluster 6 | Cluster 7 |
|---|---|---|---|---|---|---|---|---|
| Urological surgery | 359 (8) | 32 (4) | 44 (7) | 38 (7) | 187 (12) | 12 (5) | 34 (5) | 12 (4) |
| Gynecologic surgery | 468 (10) | 155 (22) | 88 (14) | 57 (10) | 97 (6) | 16 (7) | 43 (6) | 12 (4) |
| Other surgery | 179 (4) | 37 (5) | 13 (2) | 16 (3) | 60 (4) | 3 (1) | 40 (6) | 10 (4) |
| ICU admission within first 24 hours, n (%) | 3192 (19) | 690 (18) | 458 (13) | 420 (15) | 491 (27) | 581 (26) | 266 (34) | 286 (16) |
| **Admission diagnostic groups** | | | | | | | | |
| **Diagnostic group type, n (%)** | | | | | | | | |
| Diseases of the circulatory system | 2968 (18) | 532 (14) | 579 (16) | 628 (22) | 230 (13) | 359 (16) | 145 (19) | 495 (28) |
| Respiratory and infectious diseases | 1520 (9) | 319 (8) | 271 (8) | 195 (7) | 50 (3) | 511 (23) | 17 (2) | 157 (9) |
| Complications of pregnancy and childbirth | 1177 (7) | 464 (12) | 354 (10) | 201 (7) | 26 (1) | 73 (3) | 18 (2) | 41 (2) |
| Diseases of the digestive and genitourinary systems | 2139 (13) | 429 (11) | 445 (12) | 399 (14) | 354 (19) | 183 (8) | 115 (15) | 214 (12) |
| Injury and poisoning | 1943 (12) | 423 (11) | 427 (12) | 323 (11) | 229 (12) | 229 (10) | 123 (16) | 189 (11) |

| Variables | Total | Cluster 1 | Cluster 2 | Cluster 3 | Cluster 4 | Cluster 5 | Cluster 6 | Cluster 7 |
|---|---|---|---|---|---|---|---|---|
| Diseases of the musculoskeletal/connective tissue and skin | 1417 (8) | 221 (6) | 256 (7) | 219 (8) | 364 (20) | 84 (4) | 156 (20) | 117 (7) |
| Neoplasms | 1074 (6) | 224 (6) | 150 (4) | 140 (5) | 338 (18) | 50 (2) | 125 (16) | 47 (3) |
| Symptoms; signs; and ill-defined conditions | 1183 (7) | 289 (8) | 311 (9) | 254 (9) | 32 (2) | 148 (7) | 15 (2) | 134 (8) |
| Diseases of the nervous system and mental illness | 1477 (9) | 375 (10) | 382 (11) | 275 (10) | 98 (5) | 155 (7) | 21 (3) | 171 (10) |
| Endocrine; nutritional; and metabolic diseases and immunity disorders | 551 (3) | 122 (3) | 140 (4) | 78 (3) | 62 (3) | 65 (3) | 12 (2) | 72 (4) |
| **Organ dysfunction within first 24 hours** | | | | | | | | |
| **Maximum SOFA score, median (25th, 75th)** | | | | | | | | |
| All patients | 2 (1, 4) | 2 (1, 4) | 1 (0, 3) | 2 (1, 4) | 4 (3, 6) | 3 (1, 4) | 5 (4, 6) | 2 (0, 4) |
| Patients in ICU | 4 (3, 7) | 5 (3, 8) | 4 (2, 6) | 4 (2, 6) | 5 (4, 7) | 4 (3, 7) | 6 (4, 9) | 4 (2, 6) |
| Patients on ward | 2 (1, 4) | 1 (1, 3) | 1 (0, 2) | 1 (1, 3) | 4 (3, 5) | 2 (1, 3) | 4 (4, 5) | 1 (0, 3) |

| Variables | Total | Cluster 1 | Cluster 2 | Cluster 3 | Cluster 4 | Cluster 5 | Cluster 6 | Cluster 7 |
|---|---|---|---|---|---|---|---|---|
| Maximum MEWS score, median (25th, 75th) | | | | | | | | |
| All patients | 2 (1, 3) | 2 (1, 3) | 2 (1, 2) | 1 (1, 2) | 2 (1, 2) | 2 (2, 4) | 2 (1, 3) | 2 (1, 3) |
| Patients in ICU | 3 (2, 4) | 3 (2, 5) | 3 (2, 4) | 2 (2, 3) | 2 (2, 3) | 4 (3, 5) | 3 (2, 4) | 3 (2, 4) |
| Patients on ward | 2 (1, 2) | 2 (1, 2) | 1 (1, 2) | 1 (1, 2) | 1 (1, 2) | 2 (1, 3) | 2 (1, 2) | 2 (1, 3) |
| **Resource utilization** | | | | | | | | |
| Admitted to ICU, n (%) | 3874 (23) | 864 (23) | 603 (17) | 552 (19) | 512 (28) | **706 (32)** | 284 (36) | 353 (20) |
| Days in ICU, median (25th, 75th) | 3 (2, 6) | 3 (2, 7) | 3 (2, 6) | 3 (2, 6) | 3 (2, 5) | 4 (2, 7) | 4 (2, 7) | 3 (2, 6) |
| Days in ICU greater than 48 hours, n (%) | 2487 (15) | 566 (15) | 379 (11) | 329 (11) | 303 (17) | **483 (22)** | **206 (26)** | 221 (13) |
| Hospital days, median (25th, 75th) | 4 (2, 7) | 4 (2, 7) | 4 (2, 6) | 3 (2, 6) | 3 (2, 5) | 5 (3, 8) | 4 (3, 7) | 4 (2, 6) |
| Mechanical Ventilation, n (%) | 5757 (34) | 991 (26) | 821 (23) | 768 (27) | 1581 (86) | 498 (23) | **707 (90)** | 391 (23) |
| Mechanical Ventilation days, median (25th, 75th) | 1 (1, 2) | 1 (1, 2) | 1 (1, 2) | 1 (1, 2) | 1 (1, 1) | 1 (1, 3) | 1 (1, 1) | 1 (1, 2) |

| Variables | Total | Cluster 1 | Cluster 2 | Cluster 3 | Cluster 4 | Cluster 5 | Cluster 6 | Cluster 7 |
|---|---|---|---|---|---|---|---|---|
| Mechanical Ventilation hours, median (25th, 75th) | 6 (3, 13) | 6 (2, 24) | 5 (2, 12) | 5 (2, 11) | 7 (4, 10) | 7 (2, 45) | 8 (5, 13) | 4 (2, 12) |
| Mechanical Ventilation greater than 2 days, n (%) | 771 (5) | 215 (6) | 128 (4) | 87 (3) | 68 (4) | 155 (7) | 53 (7) | 65 (4) |
| Administration of vasopressor or inotropes with first 24 hours, n (%) | 4316 (26) | 803 (21) | 556 (16) | 493 (17) | 1376 (75) | 245 (11) | 664 (85) | 179 (10) |
| **Complications** | | | | | | | | |
| Acute kidney injury, n (%) | 2741 (16) | 687 (18) | 499 (14) | 422 (15) | 214 (12) | **504 (23)** | 91 (12) | 324 (19) |
| Community-acquired AKI, n (%) | 1565 (9) | 430 (11) | 259 (7) | 248 (9) | 113 (6) | **311 (14)** | 41 (5) | 163 (9) |
| Venous Thromboembolism, n (%) | 937 (6) | 251 (7) | 220 (6) | 136 (5) | 57 (3) | **169 (8)** | 33 (4) | 71 (4) |
| Sepsis, n (%) | 1913 (11) | 634 (17) | 316 (9) | 194 (7) | 73 (4) | **539 (24)** | 43 (5) | 114 (7) |
| Thirty-day mortality, n (%) | 705 (4) | 217 (6) | 103 (3) | 105 (4) | 22 (1) | **189 (9)** | 12 (2) | 57 (3) |

| Variables | Total | Cluster 1 | Cluster 2 | Cluster 3 | Cluster 4 | Cluster 5 | Cluster 6 | Cluster 7 |
|---|---|---|---|---|---|---|---|---|
| Three-year mortality, n (%) | 3324 (20) | 865 (23) | 614 (17) | 517 (18) | 197 (11) | **676 (31)** | 79 (10) | 376 (22) |

Cluster 1 contained the highest proportion of female patients (62%) and patients who were admitted for complications of pregnancy and childbirth (12%) and gynecologic surgery (22%). This cluster was the youngest (median age 50 years) and had the lowest systolic blood pressure values at the time of admission.

Cluster 2 contained a high proportion of patients undergoing general surgery (35%) operations, and the lowest proportion of patients who were admitted to an ICU during their hospitalization (17%). This cluster had relatively normal vital sign values and trends within six hours of admission.

Cluster 3 was composed of a heterogeneous group of patients with minimal organ dysfunction and short lengths of stay in the hospital (median 3 days). This cluster had the lowest heart rates of any cluster at all time-points within six hours of admission.

Cluster 4 had the highest proportion of routine elective admissions (84%), usually to a surgical service (83%), and had the highest proportion of patients admitted with a primary diagnosis of neoplasm (18%) and lowest 30-day mortality (1%). Eighty-five percent of all patients in this cluster had surgery on the day of admission, consistent with observations that 86% had mechanical ventilation and approximately two hours after admission, patients in this cluster had decreasing respiratory rates and increased blood oxygen saturation, reflecting intraoperative mechanical ventilation.

Cluster 5 had the highest proportion of emergent admissions (90%) and transfers from other hospitals (22%), usually to non-surgical services (87%), with the highest proportion of patients

29with respiratory and infectious diseases (23%) and suffered the highest incidence of several complications including acute kidney injury (23%), sepsis (24%), 30-day mortality (9%), and 3-year mortality (31%). This cluster had the highest heart rates at the time of admission (around 100 beats per minute) and six hours after admission (greater than 90 beats per minute).

Cluster 6 contained a high proportion of routine elective admissions (83%), the highest proportions of patients undergoing surgery on the day of admission (89%), the highest proportion of patients admitted to an ICU transfer within 24 hours (36%) and remaining in an ICU for more than 48 hours (26%). Systolic and diastolic blood pressures and body temperatures decreased substantially during the first six hours of admission in this cohort, along with decreasing respiratory rates and increasing blood oxygen saturations, consistent with undergoing general anesthesia and surgery.

Cluster 7 had the highest median age (64 years), greatest proportion of African American patients (33%), and the highest incidence of chronic kidney disease (26%) and end-stage renal disease (11%). These patients presented with the highest systolic blood pressure values, which remained persistently high (greater than 160 mmHg) six hours after admission.

**Discussion**

Using cluster analyses of early vital sign measurements to identify phenotypes in a heterogeneous cohort of hospitalized patients is novel. Therefore, it is difficult to compare our results with previous work. However, others have reported that clustering can identify subgroups of patients within larger cohorts of patients that have similar clinical presentations, such as sepsis and diastolic heart failure. Seymour et al.[21] performed clustering analyses on sepsis patients with the rationales that sepsis pathophysiology is heterogeneous, and identification of distinct sepsis phenotypes may facilitate provision of targeted therapies. This rationale is supported by the failure of nearly all sepsis drug trials. Clustering was performed on both clinical and immune response biomarker variables, which identified four distinct clusters. Simulations were performed in which varying proportions of each cluster were applied to previously reported randomized controlled trials. These simulations suggested unique treatment responses for different clusters. Similarly, Shah et al.[27] performed clustering analyses on patients with heart failure and preserved ejection fraction, using electrocardiogram and echocardiogram data as well as clinical variables for clustering. This study identified three distinct phenotypes that had unique clinical outcomes, even while adjusting for traditional risk factors. These findings suggest that clustering methods can identify phenotypic subgroups of patients that are not identifiable by traditional clinical parameters, and may have different treatment responses and clinical outcomes. Our study applies clustering methods to any hospitalized patient, thereby identifying more broad, generalized patterns relating to overall patient acuity and trajectory rather than targeted treatment responses for patients with established diagnoses.

Our study was limited by using data from a single institution, limiting the generalizability of our findings. Although we used retrospective data, it seems unlikely that selection bias significantly affected results, because all adult patients admitted to the hospital for longer than six hours were included. In this study, input features were constrained to the first six hours following hospital admission so that phenotypes could be identified early after hospital admission. However, it is possible that the same advantage could be achieved while incorporating historical patient data from previous encounters in the electronic health record. Further research would be necessary to determine whether incorporating historical patient data is advantageous. Finally, the ability of early clustering to augment clinical prognostication and decision-making remains theoretical until it is evaluated in a prospective clinical trial.

**Conclusions**

In this paper, we propose a novel deep interpolation network to extract the latent representations from the sparse and irregularly sampled time-series vital signs measured within six hours of hospital admission, and based on the extracted features, the clustering analyses identified seven distinct patient phenotypes. These clusters had unique pathophysiological signatures and clinical outcomes, and did not simply recapitulate known, recognized clinical phenotypes. Identifying patient phenotypes during the early stages of hospital admission may have important implications for clinical decision-support under time constraints and uncertainty. Beyond simple mortality predictions, cluster analyses can potentially elucidate disease etiology. Future research should seek external validation of these findings and investigate the utility of incorporating historical patient data from previous encounters in the electronic health record. It remains unknown whether early identification of patient phenotypes leads to improved clinical decision-making and outcomes.